# Enhancing Deep Learning with Optimized Gradient Descent: Bridging Numerical Methods and Neural Network Training

1st Yuhan Ma
Johns Hopkins University
Baltimore, USA

2nd Dan Sun
Washington University in St. Louis
St. Louis, USA

3rd Erdi Gao
New York University
New York, USA

4th Ningjing Sang
Columbia University
New York, USA

5th Iris Li
New York University
New York, USA

6th Guanming Huang*
University of Chicago
Chicago, USA

*Abstract*—Optimization theory serves as a pivotal scientific instrument for achieving optimal system performance, with its origins in economic applications to identify the best investment strategies for maximizing benefits. Over the centuries, from the geometric inquiries of ancient Greece to the calculus contributions by Newton and Leibniz, optimization theory has significantly advanced. The persistent work of scientists like Lagrange, Cauchy, and von Neumann has fortified its progress. The modern era has seen an unprecedented expansion of optimization theory applications, particularly with the growth of computer science, enabling more sophisticated computational practices and widespread utilization across engineering, decision analysis, and operations research. This paper delves into the profound relationship between optimization theory and deep learning, highlighting the omnipresence of optimization problems in the latter. We explore the gradient descent algorithm and its variants, which are the cornerstone of optimizing neural networks. The chapter introduces an enhancement to the SGD optimizer, drawing inspiration from numerical optimization methods, aiming to enhance interpretability and accuracy. Our experiments on diverse deep learning tasks substantiate the improved algorithm's efficacy. The paper concludes by emphasizing the continuous development of optimization theory and its expanding role in solving intricate problems, enhancing computational capabilities, and informing better policy decisions.

## I. INTRODUCTION

Optimization theory has long been pivotal in advancing both mathematical and computational sciences. Deeply embedded in economic theory [1-3], it has expanded across various domains, including medical diagnostics [4-6], computer vision [7-9], and text analysis [10-13], driven by the need to solve increasingly complex problems. The evolution of this field has been marked by significant milestones—from the geometric explorations of ancient Greece to the calculus innovations by Newton and Leibniz, and further, through the contributions of Lagrange, Cauchy, and von Neumann. Optimization theory has always been an important scientific tool in the fields of mathematics and computer science research from ancient times to the present. It has been continuously developing and has made significant contributions to various other fields. Now, it is being used to solve a variety of complex problems, enhance computer performance, and help policymakers make better decisions. These historical advancements laid the groundwork for the modern applications of optimization theory, which now encompass diverse areas such as neural network[14-16], deep learning[17-20] , and big data analytics.

Optimization problems are ubiquitous in deep learning. Taking regression problems as an example, we hope that the final output of the deep learning model can be as close to the true value as possible, which is actually solving an optimization problem. In addition, we can see the shadow of optimization problems in classification and object detection problems. So, can we use optimization theory to help neural networks solve optimization problems? In fact, the gradient descent algorithm in deep learning is closely related to optimization theory, and improvements to optimizers such as SGD-M [21], Adagrad [22], and RMSprop [23] are all based on improvements to optimization theory. The focus of the aforementioned improvement methods is to accelerate the training speed of the optimizer, while this chapter aims to improve SGD from the perspective of interpretability and accuracy enhancement.

This paper first briefly describes the connection between SGD [24] and the forward Euler method in numerical optimization methods. Based on this, we have improved the SGD optimizer, and have verified the good performance of the improved algorithm on various types of deep learning tasks. The experimental results show that improving the optimizer with higher-order numerical differentiation is feasible.

## II. BACKGROUND

### A. Training Process of Deep Neural Networks

The training of deep neural networks can be viewed as a non-convex optimization problem, as shown below:

$$\min_{W \in \mathbb{R}^n} L(W) = \min_{W \in \mathbb{R}^n} \frac{1}{|D|} \sum_{x \in D} L(x; W) \quad (1)$$

where $L(W)$ is the loss function; $W$ represents the model parameters; $D$ represents the dataset; $x$ represents the input to the model.

### B. Gradient Descent and Stochastic Gradient Descent

The goal of training neural networks is to find a set of appropriate parameters $w$ to minimize $L$, which is an optimization problem. Typically, we can use the Gradient Descent (GD) method to solve such problems. The iterative process of GD can be written as

$$W_{k+1} = W_k - \varepsilon \nabla_{W_k} L(W_k) \quad (2)$$

where $\varepsilon$ represents the training step size, and $k$ represents the iteration number. Since the GD method calculates the gradient by using all input samples at once, training deep neural networks with the GD method can take a long time. To improve the training efficiency of deep neural networks, Stochastic Gradient Descent (SGD) is often used instead of GD. The iterative process of SGD can be derived as

$$W_{k+1} = W_k - \varepsilon \nabla_{W_k} L_S(x; W) \quad (3)$$

where $L_S(x; W)$ can be specifically expressed as

$$\min_{W \in \mathbb{R}^n} L_S(x; W) = \min_{W \in \mathbb{R}^n} \frac{1}{|S|} \sum_{x \in S} L(x; W) \quad (4)$$

where $S$ is a subset of the dataset D. By observing equations, it is not difficult to find that the iterative forms of the GD algorithm and the SGD algorithm are very similar to the forward Euler method. Inspired by this discovery, this chapter re-examined the relationship between optimizers and numerical methods and attempted to improve SGD with a high-precision numerical differentiation.

## III. METHOD

### A. Gradient Descent Algorithm and Numerical Methods

This section first analyzes the connection between the gradient descent algorithm and the forward Euler method. Then, based on the connection between the two, the SGD algorithm's iterative method is improved using the Taylor multi-step difference method, and the TM-SGD algorithm is proposed.

Gradient Descent Algorithm and Numerical Methods We first analyze GD. Rewrite equation as

$$\frac{W_{k+1} - W_k}{\varepsilon} = -\nabla_{W_k} L(W_k) \quad (5)$$

Assuming that the training process of the deep neural network is time-related, the left side of equation can be considered as the derivative of WW. Subsequently, we obtain a system of ordinary differential equations (ODE) as

$$\dot{W} = -\nabla_W L(W) \quad (6)$$

Thus, we can view the iterative process of GD as solving the ODE using the forward Euler method. Observing equations, it can be found that in SGD, $\nabla_{W_k} L_S(x; W)$ is used to estimate $\nabla_{W_k} L(x; W)$ in GD. Next, we begin a qualitative analysis of the relationship between GD and SGD. Assume that $\dot{W} = -\nabla_W L(W) + \xi$ follows some unknown distribution $\xi$. Therefore, equation can be written as

$$\dot{W} = -\nabla_W L(W) + \xi \quad (7)$$

Observing the above equation, it is not difficult to find that the iterative process of SGD is actually similar to that of GD, but for SGD, its iterative process is solving the ODE with unknown distribution noise using the forward Euler method.

### B. Taylor Multi-step Method

Based on the Taylor series, we constructed a high-precision multi-step numerical method. Compared to the forward Euler method, the truncation error of the Taylor multi-step method is $O(\tau^3)$, while the truncation error of the forward Euler method is $O(\tau^2)$. This indicates that the Taylor multi-step method has higher computational accuracy than the forward Euler method. In Section 2.5, we conducted some numerical experiments to verify the superiority of the Taylor multi-step method.

### C. TM-SGD

Based on the analysis above, we discovered the connection between the SGD algorithm and numerical methods. When we use the Taylor multi-step method to improve the SGD algorithm, we obtain the following iterative formula:

$$W_{k+1} = 1.5 W_k - W_{k-1} + 0.5 W_{k-2} - \varepsilon \nabla_{W_k} L_S(x; W) \quad (8)$$

However, when we directly apply formula to improve SGD and propose TM-SGD, the neural network models trained by TM-SGD often perform poorly. In fact, in the numerical experiments, the constructed Taylor multi-step method exhibited significant solution fluctuations in the early stages of solving the problem. From the perspective of control theory, this is because the time lag affects the stability of the dynamic system. Note that the constructed Taylor multi-step method includes three terms (i.e., $\theta(x_k), \theta(x_{k-1}), \theta(x_{k-2})$), while the forward Euler equation only includes one term (i.e., $\theta(x_k)$). Therefore, the constructed Taylor multi-step method relies on more previous data to predict the value of the next time step than the forward Euler method.

## IV. EXPERIMENT

### A. Experiment Settings

#### 1) Dataset

This chapter conducts experiments on various tasks to verify the excellent performance of the improved SGD algorithm. Initially, the datasets involved in the experiments will be presented in this section. Subsequently, the datasets and models involved in the TM-SGD are provided in the following text.

The datasets involved in the experiments of this chapter are as follows:

- CIFAR-10/100 [25]. Benchmark datasets featuring 60,000 32x32 color images in 10/100 classes for object recognition tasks.

- VOC-2007 [26]. This dataset contains 9963 images for evaluating the model's performance in object detection and scene segmentation tasks.
- YouTube Faces [27]. A dataset of human faces extracted from YouTube videos, used for face recognition and analysis.
- Sneaker Shoes [28]. This dataset was obtained using a web crawler on the website, containing images of 7049 different types of shoes.
- R8 [29]. This dataset is a subset of the Reuters dataset, divided into 5485 training documents and 2189 test documents.
- Avazu [30]. This dataset contains 11 consecutive days of user shopping information from Avazu for testing the model's click-through rate prediction capability.

The datasets and models involved in the TM-SGD experiments are shown in Table 1. Experimental Overview: Image Classification (IC), Object Detection (OD), Segmentation (SEG), Facial Key Points Detection (FKPD), Image Generation (IG), Text Classification (TC), Click-through Rate Prediction (CTR-P). To further enhance the evaluation of the TM-SGD algorithm across the datasets listed, a sophisticated data processing method as outlined by Li et al. will be integrated into the experimental setup [31]. This method, derived from their work on creating accessibility linked data from publicly available datasets, involves advanced techniques for data integration and transformation, ensuring that the datasets are optimally prepared for complex analytical tasks. By applying the principles and methodologies, we can preprocess the datasets to improve the reliability and validity of the experimental outcomes, particularly in areas such as image and text classification, where data quality significantly influences model performance.

Table 1. Datasets

| Task | Dataset | Model |
|---|---|---|
| IC | CIFAR-10&100 | VGG & ResNet |
| OD | VOC-2007 | Fast-RCNN & RFBNet |
| SEG | VOC-2007 | Lraspp |
| FKPD | YouTube Faces | PreActResNet34 |
| IG | Sneakers Shoe | GAN |
| TC | R8 | LSTM |
| CTR-P | Avazu | Wide&Deep |

*2) Parameter*

To ensure the fairness of the experiments, the learning rate, training duration, and learning rate schedule are consistent across all experiments. Additionally, the majority of the experimental code in this chapter is based on existing GitHub projects, hence the settings for learning rate, training duration, and batch size of the data follow the original project's configuration. It should be noted that the GitHub project code used in the experiments has been simplified to some extent compared to the current state-of-the-art (SOTA) training code. For instance, for the input preprocessing of image data, there is no random flipping, random cropping, or optimal coefficient setting for normalization. This leads to some of the optimizer experimental results obtained in this paper being slightly lower than their corresponding SOTA baseline results.

As shown in 2, the training settings related to TM-SGD are provided. In particular, Switch Epoch is a hyperparameter, and this paper conducts many experiments on Switch Epoch to explore the impact of different Switch Epoch on model performance. Considering that the full experimental results would take up a lot of space, the paper first presents a simplified version of the experimental results for discussion and analysis, and places the complete experimental results after the discussion and analysis.

Table 2. Parameters

| Task | Epoch | Learning rate | Batchsize | Switch Epoch |
|---|---|---|---|---|
| IC | 10 | 0.1 | 256 | 75 |
| OD | 210 | 4e-3 | 32 | 10 |
| SEG | 30 | 1e-4 | 4 | 200 |
| FKPD | 20 | 1e-2 | 10 | 5 |
| IG | 120 | 2e-3 | 64 | 50 |
| TC | 120 | 1e-2 | 5 | 100 |
| CTR-P | 20 | 1e-3 | 256 | 15 |

*B. Experiment Results Analysis*

*1) ObjectDetection*

In Table 3, models trained with TM-SGD consistently outperform those trained with SGD. For instance, Fast-RCNN [32] trained with TM-SGD achieved a mean average precision (mAP) of 62.29%, which is better than the 61.77% mAP achieved by Fast-RCNN trained with SGD. Similar results can be observed in RFBNet. The model trained with TM-SGD reached an mAP of 77.32%, while the model trained with SGD only achieved an mAP of 77.13%. For segmentation tasks, TM-SGD also demonstrated strong performance. As shown in Table 3-6, for segmentation tasks, Lraspp trained with TM-SGD reached a mean intersection over union (mIoU) of 70.8, outperforming Lraspp trained with most other optimizers (IoU for SGD is 70.1%, for Adam is 68.6%, and for Apollo is 68.5%).

Table 3. Experiment Results

| Model | Optimizer | mAP(%) |
|---|---|---|
| Fast-RCNN | SGD | 61.77±0.02 |
|  | TM-SGD | 62.29±0.05 |
|  | SGD-M | 65.89±0.06 |
|  | Adam | - |
|  | Apollo | 55.07±0.03 |
| RBFNet | SGD | 77.13±0.04 |
|  | TM-SGD | 77.32±0.05 |
|  | SGD-M | 79.33±0.07 |
|  | Adam | - |
|  | Apollo | 74.51±0.01 |

*2) Face Detection*

As shown in Table 4, the performance of TM-SGD is superior to that of SGD and other types of optimizers (with

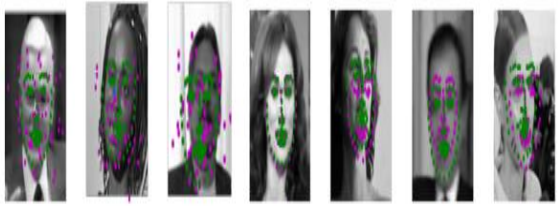

Figure1. Prediction Results

the exception of SGD-M). For instance, PreActResNet34 trained with TM-SGD achieved a test mean squared error (MSE) of 0.1241, whereas the MSE for SGD was 0.1306, for Adam it was 0.132, and for Apollo it was 0.1271. As depicted in Figure 1, we selected the prediction results of 7 test images to demonstrate the excellent performance of TM-SGD. In Figure 1, it can be observed that:

1) Models trained with SGD sometimes fail to identify facial key points, while models trained with TM-SGD can accurately identify them;

2) The facial key points predicted by models trained with TM-SGD are closer to the annotated facial key points in the test images compared to those predicted by models trained with SGD.

Table 4. Loss Results

| Optimizer | MSE | Loss |
| --- | --- | --- |
| SGD | 0.1306±6e-3 | 0.01545±5e-4 |
| TM-SGD | 0.1241±3e-3 | 0.01354±7e-4 |
| SGD-M | 0.0643±8e-4 | 0.00364±1e-4 |
| Adam | 0.1320±9e-3 | 0.01660±7e-4 |
| Apollo | 0.1271±4e-3 | 0.01513±8e-4 |

*3) Image Generation*

As shown in Figure 2, we present the results of training a GAN with different optimizers over various training periods. Commonly, the Frechet Inception Distance score (FID) is used as a quantitative metric to evaluate the performance of GANs [33]. GAN training generally relies on the Adam optimizer, and training GANs with other optimizers usually yields poorer results. In this experiment, the purpose of using Adam is not for comparison but to provide a benchmark for performance comparison. Comparing with Adam will be one of our future works. In Figures 2 (b)-(d), although it is difficult to distinguish which set of shoe generations is closer to reality, it can be clearly observed that GANs trained with TM-SGD produce a more diverse range of shoes. In contrast, GANs trained with SGD and SGD-M tend to generate white shoes. The lower the FID, the better the performance of the GAN. In this experiment, the FID for Adam is 0.0553, for SGD it is 0.0925, for SGD-M it is 0.0916, and for TM-SGD it is 0.081. This indicates that GANs trained with Adam and TM-SGD outperform those trained with SGD and SGD-M. In the task of image generation, TM-SGD significantly outperforms SGD in both image quality and quantitative metrics.

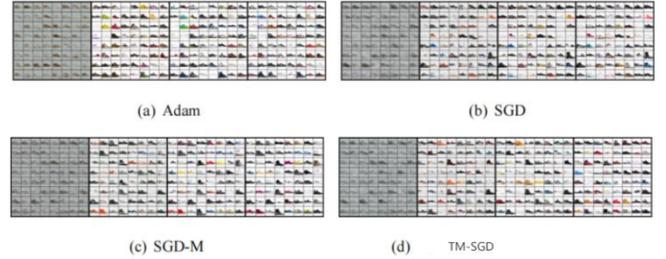

Figure2. Prediction Results

*4) Text Classification*

As shown in Table 5, the models trained with the Adam, SGD, and Apollo algorithms achieved accuracies of 90.1%, 90.6%, and 89.4%, respectively, on the R8 dataset, while the TM-SGD algorithm achieved an accuracy of 90.9%. This indicates that TM-SGD has effectively enhanced the performance of the given LSTM model on the text classification task.

Table 5. Experiment Results

| Optimizer | SGD | TM-SGD | SGD-M | Adam | Apollo |
| --- | --- | --- | --- | --- | --- |
| ACC | 90.6 | 90.9 | 90.4 | 90.1 | 89.4 |

V. CONCLUSIONS

This paper begins by elucidating the connection between GD/SGD and the forward Euler method. Subsequently, the numerical experimental results demonstrate that the Taylor multi-step method has higher computational accuracy than the forward Euler method. Considering the impact of computational accuracy on solving ODE problems, this chapter improves the SGD algorithm using the Taylor multi-step difference method and proposes TM-SGD. Thereafter, we tested the performance of the TM-SGD algorithm in various different task scenarios. The experimental results show that the performance of the improved TM-SGD is superior to that of the original SGD. Additionally, taking SGD-M as an example, we improved it using the Taylor multi-step method and discussed the performance enhancement and existing shortcomings of the improved TM-SGD-M. The work in this chapter indicates that using numerical difference methods to improve optimizers is a straightforward approach to optimizer enhancement. Compared to other optimizer improvement methods, the optimizer proposed in this paper is easy to modify the code, and has certain interpretability and extensibility, which means that the method proposed in this paper does not require changing the original training method and hyperparameter settings of the optimizer to enhance the performance of the improved method. The application of these sophisticated numerical techniques in optimization algorithms opens new avenues for industries such as healthcare, where predictive modeling and diagnostic accuracy are paramount; finance, where complex risk assessments and algorithmic trading benefit from enhanced computational capabilities; and autonomous vehicle navigation, which relies on precise and efficient real-time decision-making systems. The continued exploration and adaptation of such methods promise significant strides in the field of computational intelligence, potentially revolutionizing how industries leverage AI to solve complex, real-world problems.